\documentclass{article}
\usepackage[T1]{fontenc}
\usepackage[utf8]{inputenc}
\usepackage{color}
\usepackage{float}
\usepackage{textcomp}
\usepackage{amstext}
\usepackage{amssymb}
\usepackage{graphicx}
\usepackage[unicode=true,
 bookmarks=false,
 breaklinks=false,pdfborder={0 0 1},backref=section,colorlinks=false]
 {hyperref}

\makeatletter

\providecommand{\tabularnewline}{\\}
\floatstyle{ruled}
\newfloat{algorithm}{tbp}{loa}
\providecommand{\algorithmname}{Algorithm}
\floatname{algorithm}{\protect\algorithmname}

\usepackage{url}
\usepackage{booktabs}
\usepackage{amsfonts}
\usepackage{nicefrac}
\usepackage{microtype}
{}
\usepackage{textcomp}
\usepackage{amstext}
\usepackage{algorithm}
\usepackage{algpseudocode}
\usepackage{textgreek}
\usepackage{todonotes}

\title{SCSP: Spectral Clustering Filter Pruning with Soft Self-adaption Manners}

\author{Huiyuan Zhuo, Xuelin Qian, Yanwei Fu, Heng Yang, Xiangyang Xue
}

\makeatother

\begin{document}

\maketitle 
\begin{abstract}
Deep Convolutional Neural Networks (CNN) has achieved significant
success in computer vision field. However, the high computational cost of the deep complex models prevents the deployment on edge devices with limited memory and computational resource. In this paper, we proposed
a novel filter pruning for convolutional neural networks compression, namely spectral clustering filter pruning with
soft self-adaption manners (SCSP). We first apply spectral clustering
on filters layer by layer to explore their intrinsic connections and
only count on efficient groups. By self-adaption manners, the pruning
operations can be done in few epochs to let the network gradually
choose meaningful groups. According to this strategy, we not only
achieve model compression while keep considerable performance, but also find a novel angle to interpret the model
compression process. 
\end{abstract}

\section{Introduction}

In recent years, Deep Neural Networks (DNNs) have achieved significant
progress in the most of computer vision tasks, e.g. classification
\cite{KrizhevskySH12} and object detection \cite{overfeat}. However,
with the boost of accuracy performance, the architecture of networks becomes
deeper and wider, which leads to much higher requirement of computational resource for inference. 
For example, the popular network ResNet-152 \cite{resnet}
has 60.2 million parameters with 232 MB model size, and it needs
11.3 billion float point operations for one forward process. Such
huge cost of resource make it hard to deploy a typical deep model
on resource constrained devices, \emph{\emph{e.g.},} mobile phones or embedded
gadgets. With the desire of deployment for edge computing, the study of deep
neural network compression has obtained considerable attention on
both academia and industry.

In the 1990s, LeCun \emph{et al.} \cite{lecun1990optimal} first observed
that if some weights have less influence on final decision, then we
could prune them with slight accuracy loss. Pruning is one of the most popular models for model compression, since it will
not hurt the original network structure. Recently, a lot of works
based on pruning has been published. Han \emph{et al.} \cite{weight_connects}
pruned non-important connections for model compression (weights pruning).
He \emph{et al.} \cite{channel_pruning} and Luo et al. \cite{thinet}
both utilized filter pruning. The former applied two-step algorithms
to realize channel pruning for deep neural network acceleration, and
the latter proposed a novel pruning mechanism that whether a filter
can be pruned depends on the outputs of its next layer.

Intuitively, the filters of the convolutional neural networks are
not independent. Despite that different filters try
to learn features from different points, some points may have potential
similarity or connections. Therefore, we believe that filters in networks
could be divided into several groups, namely
clustering. Contrary to the filter pruning method in \cite{yoon2017combined}
which considered each filter as a group, we propose a novel spectral
clustering filter pruning for CNNs Compression with Soft Self-adaption Manners. First, we cluster filters into $k$ groups with spectral
clustering, then we rank the groups by their contributions (weights)
to the final decision. Some filers in a group will be pruned if they
have low influences. Furthermore, inspired by \cite{soft_filtering},
we combine our proposed method with soft filter pruning, where the
pruned filters will be updated iteratively during training to retain
the accuracy.

\noindent \textbf{{} Contribution.}
Our contributions of this paper are as follows. (1) Considering
the connections in filters, we proposed a novel spectral clustering
filter pruning for CNNs Compression, which pruned a group of related
but redundant filters instead of each independent filters. (2) Our
proposed method provided maximum protection of correlations between
filters with spectral clustering, therefore our approach can train a model
from scratch with fast convergence, meanwhile, the model can achieve
comparable performance with much fewer parameters. (3) Experiments on
two datasets (\emph{i.e.}, MNIST and Cifar-10) demonstrate
the high effectiveness of our proposed approach. 

\section{Related Works}

\subsection{Spectral Clustering}

Spectral clustering algorithms attempted to partition data points
into groups such that the points of a group are similar to each other
and dissimilar to others outside of the group. Its essence is point
to point clustering method that converts clustering problem into graph
optimal partition problem. Comparing with other traditional clustering
algorithms (\emph{e.g.}, k-means and mixture models ), it enables better
results and faster in convergence. Given some data points ${x}_{1},...,{x}_{n}$,
the basic spectral clustering algorithms can be formulated as below:
\begin{enumerate}
\item Calculate the degree matrix $D$ and the adjacency matrix $A$;
\item Calculate the Laplacian matrix $L=D-A$;
\item Calculate the first $k$ eigenvectors (corresponding to the first
$k$ eigenvalues) of the Laplacian matrix $L$ with eigenvalue decomposition,
and reconstruct them as a $n\times k$ Eigen matrices;
\item Clustering the $n\times k$ Eigen matrix using k-means by row.
\end{enumerate}
Shi \emph{et al.} \cite{shi2000normalized} firstly applied spectral
clustering algorithm in images segmentation task. They formulated
the problem from a graph theoretic perspective and introduced the
normalized cut to segment the image. After that, lots of works \cite{hershey2016deep,hou2016towards,ozertem2008mean}
based on spectral clustering started to show its promising ability
in computer vision. Yang et al \cite{spectral_optical} presented
a novel kernel fuzzy similarity measure, which uses membership distribution
in partition matrix obtained by KFCM to reduce the sensitivity of
scaling parameter, for image segmentation. To overcome the common
challenge of scalability in image segmentation, Tung \emph{et al.} \cite{tung2010enabling} proposed a method which first performs
an over-segmentation of the image at the pixel level using spectral
clustering, and then merges the segments using a combination of stochastic
ensemble consensus and a second round of spectral clustering at the
segment level.

\subsection{Model Compression and Acceleration}

The previous approaches of model compression and acceleration can
be roughly divided into three parts \cite{cheng2017survey}: parameter
pruning, low-rank factorization and knowledge distillation. The parameters
of convolution or fully-connected layer in a typical CNN may have
large redundancy. Based on this situation, some works \cite{rigamonti2013learning,denton2014exploiting,jaderberg2014speeding,tai2015convolutional} concentrated on decomposing matrices/tensors to estimate the informative
parameters with \emph{low-rank factorization}. Other studies \cite{romero2014fitnets,balan2015bayesian,luo2016face}
aimed to compress models by training a new compact neural network
which could be on par with a large model. For example, \cite{hinton2015distilling}
followed a student-teacher network to realize the compact network
training, \emph{i.e.} \emph{knowledge distillation}. As the starting point
of our work, \emph{parameter pruning} 
\cite{heyangijcai,jingzhongijcai,clipq,shaohuiijcai,guiyingijcai} 
focused on removing some weights or filters which are redundant or
not important to the final performance. Molchanov et al. \cite{molchanov2016pruning}
introduced a Taylor expansion to approximate the change in the cost
function induced by removing unnecessary netowrk parameters. Li et
al. \cite{li2016pruning} presented an acceleration method to
prune filters with their connecting feature maps from CNNs that have
a small contribution to ouput. Yu et al. \cite{yu2017nisp}
applyed feature ranking techniques to measure the importance of responses
in the second-to-last layer, and propagated the importance of neurons
backwards to earlier layers. The nodes with low propagated improtance
are then pruned. Different from Hard filter pruning, He et al. \cite{soft_filtering}
proposed soft filter pruning, which allows the pruned filters to be
updated during the training procedure. Liu et al. \cite{liu2017learning}
imposed L1 regularization on the scaling factors in batch normalization
(BN) layers to identify insignificant channels (or neurons), which
are pruned at the following step.

In contrast to these existing works, we applied spectral
clustering algorithms to model compression. The intuitional idea is
that filters in networks are not independent, they could be divided into several groups (\emph{e.g.}, spectral clustering) based on potential similarity or connections.

\begin{figure}
\centering{}\includegraphics[scale=0.4]{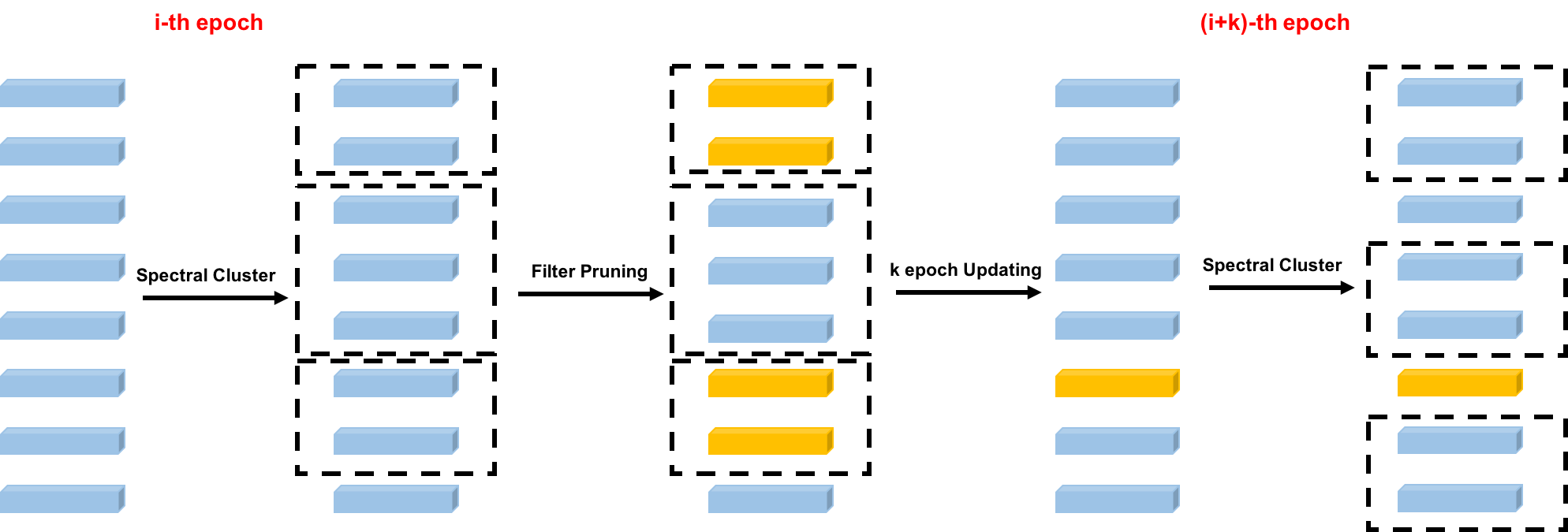}\caption{\label{fig:Overview}SCSP and training process. In $i_{th}$ epoch, we perform SCSP, firstly, do spectral clustering to group the filters. In above picture, we cluster filters into 3 groups, then do filter pruning. In next $k$ epoch, we follow normal updating process, then repeat SCSP again in $(i+k)_{th}$ epoch.}
\end{figure}

\section{Methodology}

The schematic overview of our framework has been summarized in Alg.1
and Alg.2. We will introduce the annotations and assumptions before
going into more details about how to do spectral clustering on network
filters, and how to train the model with soft self-adaption manners.

\subsection{Preliminary}

Deep neural networks can be parameterized with trainable variables.
We denote the ConvNets model as $\Omega={\left\{ \Phi^{i}\in\mathbb{R}^{H_{i}\text{\texttimes}W_{i}\text{\texttimes}I_{i}\text{\texttimes}O_{i}},i\in[1,L]\right\} }$,
where $\Phi^{i}$ is the connecting weights of the i-th convolutional
layer which connect the layer i and layer i+1. $H_{i},W_{i},I_{i}$
and $O_{i}$ are height, width, in-channels which are the input of
the $i_{th}$ layer, out-channels which are the input of the $(i+1)_{th}$
layer, respectively. L is the depth of convolutional network. This
formula can be easily tuned for fully connected layers, so we skip
the details. In order to do the spectral clustering on the filters,
we need to reshape the filters $\Phi^{i}$ to a new weight matrix
$\mathcal{M}^{i}$ with dimension $[H_{i}\times W_{i}\times I_{i},O_{i}]$,
where $O_{i}$ is the number of filters, $H_{i}\times W_{i}\times I_{i}$
is the number of weights( attributes) of single filter.

For some parameters used in our strategy, we denote bandwidth parameter
as $\varpi_{i}$ used in radial basis function, the pruning rate as
$\mathcal{P}_{i}$ used in i-th layer and the numbers of clustering
groups as $\mathcal{N}_{i}$. The reason we can do the filter clustering
into various groups and pruning is based on the assumptions that every
filter learns different aspects of the target, while some filters
are trivial for target, because the ``Black Box” of neural networks
brings some repeated efforts inadvertently.

\subsection{Spectral Clustering Filter Pruning with Soft Self-adaption manners}

In previous filter pruning for model compression, once the filters
are pruned, it will not update it again, which means the model capacities
will drop dramatically. Whereas, in our frame work, we will update
the pruned filters iteratively to retain the accuracy at same time.
Meanwhile, other model compression frameworks always reformulate the
form of objective function by combining the L1 norm and L2 norm to
get compact model. In \cite{yoon2017combined}, according to novel objective function, it
will get exclusive sparsity and group sparsity, while the definitions
about clustering groups of filters are too intuitive. To this end,
we will propose a reasonable and reliable manners to get the groups
of filters.

\subsubsection{Spectral Clustering for Filter Pruning.}

Spectral clustering has become one of the most popular modern clustering
algorithms in recent years. Compared to the traditional algorithms,
such as K-means, single linage, spectral clustering has many fundamental
advantages. When it comes to filter pruning, it's necessary to modify
some default settings.

\textbf{Cosine similarity matrix.} When analyzing the big data, especially
with hundreds of attributes, Euclidean distance suffers from the `curse
of dimensionality'. Specifically, using Euclidean distance will produce
tremendous distance when two vectors are very similar by all attributes
except for one attribute. Hence, cosine distance is recommended in
case of deep neural networks filter pruning. As summarized in algorithm
1, firstly, we compute the pair-wise cosine similarities of filters
from the same layer to form a similarity matrix $\mathcal{S}^{i}$.
To ensure the correctness, we need to eliminate the zero filters and
need some trick to avoid infinity. Without confusion, we still use
$\mathcal{M}^{i}$ to denote the filters matrix participated in pruning
process. Specifically,

\begin{equation}
\mathcal{S}^{i}\in\mathbb{R}^{O_{i}\times O_{i}}=\{\mathcal{C}_{j,k}^{i}\vert1-\frac{\mathcal{M}_{j}^{i}\mathcal{M}_{k}^{i}}{\left\Vert \mathcal{M}_{k}^{i}\right\Vert \left\Vert \mathcal{M}_{k}^{i}\right\Vert }\}
\end{equation}

\begin{equation}
\Gamma^{i}\in\mathbb{R}^{O_{i}\times O_{i}}=\{w_{j,k}^{i}\vert e^{-\left(\mathcal{C}_{j,k}^{i}\right)^{2}/\varpi_{i}}\}
\end{equation}

where $\mathcal{C}_{j,k}^{i}$ is the cosine similarity between two
filters j, k in i-th layer, using 1 to minus the cosine distance is
to fit the nature definitions of similarity that same filters' distance
is 0, more different filters has larger value. $\mathcal{M}_{k}^{i}$is
the k-th filter of i-th layer's reshaped filter matrix $\mathcal{M}^{i}$.
We implement radius basis transformation in cosine similarity traditionally,
denoted by $w_{j,k}^{i}$. We define the transformed matrix as $\Gamma^{i}$,
called ad.

\textbf{K-Means on feature matrix.} Then, we can compose the normalized
Laplacian matrix $\Pi^{i}$ based on $\Gamma^{i}$. Specifically,
we compute a diagonal matrix $\mathcal{D}^{i}$ which is a symmetric
matrix where the diagonal elements is sum of the each column of $\Gamma^{i}$.
Using $\mathcal{D}^{i}$ minus $\Gamma^{i}$ to compute the unnormalized
Laplacian matrix. Finally, we exploit the Eigen decomposition of normalized
Laplacian matrix $\Pi^{i}$ to get feature matrix $\mathcal{F}^{i}$
for clustering. To reduce the dimension of $\mathcal{F}^{i}$, we
will choose k eigenvectors corresponding to k largest eigenvalues
to compose a new feature matrix, while without confusion, we still
using $\mathcal{F}^{i}$ to illustrate the process. While in practice
we often abandon the dimension reducing operation to maintain all
information of filters. whether to perform the dimension reducing
operations or not depends on your tasks. Specifically,

\begin{equation}
\mathcal{D}^{i}\in\mathbb{R}^{O_{i}\times O_{i}}=\left(\begin{array}{ccc}
d_{1} & \ldots & \ldots\\
\ldots & d_{2} & \ldots\\
\vdots & \vdots & \ddots\\
\ldots & \ldots & d_{m}
\end{array}\right),where\ d_{m}=\sum_{n}\Gamma_{mn}^{i}
\end{equation}

\begin{equation}
\Pi^{i}\in\mathbb{R}^{O_{i}\times O_{i}}={\mathcal{D}^{i}}^{-1/2}(\mathcal{D}^{i}-\Gamma^{i}){\mathcal{D}^{i}}^{1/2}
\end{equation}

We then apply K-Means algorithm to feature matrix $\Pi^{i}$ to get
the filters group labels, denoted as $\pi^{i}\in\mathbb{R}^{O_{i}}$,
where each element in $\pi^{i}$ belongs to {[}1,2,..., $\mathcal{N}_{i}${]}.
We need to minimize the following objective function: 
\begin{equation}
\mathop{\arg\min}_{\pi^{i}}\ \ \sum_{n}^{\mathcal{N}_{i}}\sum_{f_{k}^{i}\in\pi^{i}}\Vert f_{k}^{i}-\mu_{k}\Vert_{2}^{2}
\end{equation}
where $f_{k}^{i}$ $\in\mathbb{R}^{H_{i}\times W_{i}\times I_{i}}$
is $k_{th}$ column of $\mathcal{F}^{i}$ and $\mu_{k}$ $\in\mathbb{R}^{H_{i}\times W_{i}\times I_{i}}$
is a cluster center, k $\in$ {[}1,2,..., $\mathcal{N}_{i}${]} 
\begin{algorithm}
\caption{Spectral Clustering for Filter Pruning}

function: Spectral-Clustering($\Phi^{i}$) returns the clustering
labels $\pi^{i}$

\ \ \ \ \ Inputs: filters $\Phi^{i}$, bandwidth $\varpi_{i}$,
cluster groups $\mathcal{N}_{i}$

\ \ \ \ \ Reshape the $\Phi^{i}$ to $\mathcal{M}^{i}$ with
dimension $[H_{i}\times W_{i}\times I_{i},O_{i}]$

\ \ \ \ \ Drop zero weight filter to get new filters $U_{i}$

For filter $j,$ filter $k$ in $\mathcal{M}^{i}$ do

\ \ \ \ \ \ \ \ $\mathcal{C}_{j,k}^{i}=Cosine-Distance(\mathcal{M}_{j}^{i},\mathcal{M}{}_{k}^{i})$

\ \ \ \ \ \ \ \ $w_{j,k}^{i}=RBF(\mathcal{C}_{j,k}^{i})$

Compute normalized Laplacian matrix $\Pi^{i}$

Do Eigen decomposition on $\Pi^{i}$

If reduce-dimension=False

\ \ \ \ \ \ \ \ then Return ${\mathcal{F}}^{i}$

Do K-means( ${\mathcal{F}}^{i}$ ) according to cluster groups $\mathcal{N}_{i}$

Return clustering labels $\pi^{i}$ 
\end{algorithm}

\subsubsection{Filter Pruning with Soft Self-adaption Manners}

As we mentioned before, the filter pruning operations with soft manners
ensure the capacity of model while significantly increase the computation
complexity and speed up the training. In this section, we will go
into details about how to apply soft manners to our filter pruning.

\textbf{Filter selection} After spectral clustering operations on
$\mathcal{M}^{i}$, we will get the group labels of each filter in
$i_{th}$ layer, naturally separate the filter into $\mathcal{N}_{i}$
groups. According to $\pi^{i}$, we get group set as $\mathcal{G}^{i}={\left\{ g_{k}^{i}|k\in[1,\mathcal{N}_{i}]\right\} }$,
where $g_{k}^{i}$ contains some filters with group label k.We will
evaluate the importance of each group by applying $L_{p}$ norm. In
intuition, the convolutional results of the filter with larger $L_{p}$
norm will lead to relative more activation values, thus have more
numerical impact on the target. To this end, we define the $k_{th}$
group effect in layer $i$ to indicates the total efforts of a group
to final target, denoted as ${E}_{k}^{i}$. After comparing the group
effects among different groups, we choose $\mathcal{N}_{i}\mathcal{P}_{i}$
groups to pruning, where $\mathcal{P}_{i}$ is the pruning rate in
i-th layer related with the amounts of group. Empirically, we will
use L2 norm. Like,

\begin{equation}
{E}_{k}^{i}=\sqrt[p]{\sum_{n=1}^{H_{i}\text{\texttimes}W_{i}\text{\texttimes}I_{i}}\left|m_{k}^{i}(n)\right|^{p}}
\end{equation}

where $m_{k}^{i}(n)$ is $n$-th weight of $k$-th filter in layer
$i$.

\textbf{Filter Pruning.} In filter selection above, we rank the group
effect, and choose some group to be pruned together. In practice,
we actually choose same pruning rate, whereas, for different $\mathcal{N}_{i}$,
we still prune different numbers of filters. The reason why we choose
different cluster group amounts for different layers is because different
layers has great bias on filter amounts, which means different numbers
of features will be learned. The heuristic way we choose is to choose
the group number proportional to the number of class in classification
task. On the other hand, we treated convolutional layers and fully
connected layers differently, such as, in some classification task,
we won't prune the last layers, where each filter represent all learned
features for one class. The choice of cluster groups number and the
different treatment of layers depends on task.

By now, why can we simply prune filters at same time? Intuitively,
due to the large model capacity and we will update the model in next
$t$ epoch. Finally, the model become more flexible, better generalization
and retain same accuracy.

\textbf{Reconstruction.} After pruning filters, we recover the filters
in next $t$ epoch. However, how to choose the frequency of pruning.
Specifically, how many epochs are appropriate between every two pruning
operations, which I call it `pruning gaps'? if we choose a small
pruning gaps, which will update frequently while the filters don't
have enough time to recover. Consequently, it will prune almost same
filters in following pruning steps. When it comes to large pruning
gaps, although leaving enough time for filters to recover, the model
isn't compact as expected. Hence, there is a balance between accuracy
and compactness. Empirically, set pruning gaps equals one or two epoch
is enough, meanwhile, you can set two epochs to recover before ending
the training process.

\begin{algorithm}
\caption{Filter Pruning with Soft Self-adaption Manners}

Input: model parameters $\Omega=\{\Phi^{i}\in\mathbb{R}^{H_{i}\text{\texttimes}W_{i}\text{\texttimes}I_{i}\text{\texttimes}O_{i}},i\in[1,L]\}$,
bandwidth $\varpi_{i}$, cluster groups $\mathcal{N}_{i}$ , pruning
rate $\mathcal{P}_{i}$ , pruning gaps $t$,$epoch_{max}$

Initialize $\Phi^{i}$

while $epoch\le epoch_{max}$ and $epoch\ \%\ $t$==0$ \ do

\ \ \ \ \ \ \ \ for each layer $l\in L$, do

\ \ \ \ \ \ \ \ \ \ \ \ \ \ \ \ clustering label $\pi^{i}=Spectral-Clustering(\Phi^{i})$

\ \ \ \ \ \ \ \ \ \ \ \ \ \ \ \ computing group effect
${E}_{k}^{i}$ using $L_{p}$ norm

\ \ \ \ \ \ \ \ \ \ \ \ \ \ \ \ then rank

\ \ \ \ \ \ \ \ \ \ \ \ \ \ \ \ zeroize $\mathcal{N}_{i}\mathcal{P}_{i}$
groups of filters

\ \ \ \ \ \ \ \ end for

\ \ \ \ \ \ \ \ update filters $\Phi^{i}$

end while 
\end{algorithm}

\section{Experiments}

In this section, we evaluate our proposed approach on several classification
datasets with some popular base networks.

\subsection{Datasets and Settings}

\noindent \textbf{MNIST \cite{lecun1998gradient}.} It is a classical handwritten digits dataset
which contains $70,000$ grayscale images, has a training set of 60,000
examples, and a test set of $10,000$ examples. The 5-layers convolution
neural network (LeNet-5) for handwritten digits classification opened
a new gate for deep learning.

\noindent \textbf{CIFAR-10 \cite{torralba80M}.} It's a widely-used
object classification dataset in computer vision. The dataset consists
of $60,000$, $32\times32$ color images in 10 classes (\emph{e.g.}, cat,
automobile, airplane, etc.), with $6,000$ images per class. The dataset
is divided into five training batches and one test batch, each with
$10,000$ images. The test batch contains exactly $1,000$ randomly-selected
images from each class. The training batches contain the remaining
images in random order, about $5,000$ images from each class.

\noindent \textbf{Implementation details.} Our models are implemented
on Tensorflow framework. In particular, \emph{For MNIST dataset},
we adopt two convolutional layers follows by two fully connected layers,
named LeNet-4, as the base network, then we train the network for
20 epochs with constant learning rate of 0.07. We evaluate our proposed
method with five different pruning ratios, \emph{i.e.}, $10\%$, $20\%$
,$30\%$ and $40\%$. Besides, the number of spectral clustering groups
is related to the number of filters which participate in the operation
of filter pruning. Note that we don't prune the last layer in this
paper. \emph{For CIFAR-10 setting}, we follow the LeNet-5 and ResNet-32
as the base network structures. All models are trained for 200 epochs
with multi-step learning rate policy (0.003 for the first 75 epochs,
0.0005 for the following 75 epochs, and 0.00025 for the rest epochs).
Because of compact and simple network structure, we just prune the
second the third layers in LeNet-5, however, we do filter pruning
for all convolution layers in ResNet-32.

\subsection{Evaluation Protocol}

\noindent \textbf{FLOPs.}
Floating-point operations(FLOPs) is a widely-used measure to test
the model computational complexity. We follow the criterion in \cite{molchanov2016pruning},
the formulation is described as below.

\emph{For convolutional layers}:

\begin{equation}
FLOPs=2HW\left({C}_{in}{K}^{2}+1\right){C}_{out},\label{eq:protocl_1}
\end{equation}

where $H$, $W$ are the size of input feature map, $K$ is the size
of the kernel, ${C}_{in}$ and ${C}_{out}$ are the input and output
channels respectively.

\emph{For fully connected layers}:

\begin{equation}
FLOPs=\left(2{C}_{in}-1\right){C}_{out},\label{eq:protocl_2}
\end{equation}

where ${C}_{in}$ and ${C}_{out}$ are the input and output channels
respectively.

\noindent \textbf{Parameters sparsity.}
Parameters sparsity is a percentage of how many parameters are realistically
used over all theoretic parameters, as in the following formulation:

\begin{equation}
Sparsity=\frac{\sum_{k=1}^{|{W}^{l}|}{I\left({w}_{k}^{l}\ne0\right)}}{|{W}^{l}|},\label{eq:protocl_3}
\end{equation}

where $|{W}^{l}|$ denotes the the number of all parameters in $l-th$
layer.

\subsection{Results on MNIST}

As we can see in Table \ref{tab:Results-about-MNIST}, we evaluated
our approach with five different pruning ratios, and also compared
with other state-of-art methods, \emph{e.g.}, CGES \cite{yoon2017combined}. First, with the increase of pruning ratio,
the filters become more sparse, fewer and fewer parameters can be
utilized for final decision, however, our approach still could achieve
considerable performance $98.54\%$, $98.63\%$, $98.4\%$, $98.46\%$
and $98.42\%$ with $0.1$, $0.2$, $0.3$, $0.4$, and $0.5$ pruning
ratio respectively. Second, interestingly, we notice that when the
pruning ratio is $0.2$ (or $0.1$) which means that in each training
step, we pruned $20\%$ of all groups of filters, our model with $92.43\%$
remaining parameters even could achieve better accuracy than baseline
with $0.13\%$ margin. It confirmed our intuition that in spite of
plenty of parameters in networks, they can be grouped based on relevance
and complementarity. Pruning some redundant groups (filters) leverage
others to focus on their responsibility and learn more discriminative
features. Third, comparing with CGES, our model has better performance
and lower FLOPs. Although our parameter sparsity is not as good as
CGES, we have a better balance between accuracy and computation complexity.

\subsection{Results on CIFAR-10}

\noindent \textbf{LeNet-5 base network.} We first evaluate our method
on CIFAR-10 dataset with LeNet-5, and set the pruning ratio by 0.4.
Table \ref{tab:On-LeNet-5} explains the comparison between our performance
and CGES. As we can see, our approach achieved the state-of-art performance.
After training, we pruned about $18\%$ parameters of network and
reduced $6.42\%$ FLOPs, however, our model do not loss any capacity,
still maintain the final accuracy with $71.09\%$. When looking at
the performance of CGES, it sacrificed $1.04\%$ accuracy only for
decreasing $3.6\%$ parameters. That' all thanks to the effectiveness
of Spectral Clustering Filter Pruning.

\noindent \textbf{ResNet-32 base network.} We also apply our proposed pruning method on a deeper network to show it's strength. As a deep and powerful network, ResNet is our primary choice. We utilize ResNet-32 as our base network, and the pruning ratio is set by 0.4. Results are illustrated in Table \ref{tab:On-ResNet-32}. Comparing with baseline, $12.10\%$ parameters are pruned by our approach, and we reduce the computation complexity by $11.3\%$ in FLOPs. Meanwhile, the model only loss the precision with a small margin.

\begin{table}
\centering{}%
\begin{tabular}{c|c|c|c|c|c}
\hline 
Pruned ratio  & Baseline Acc(\%)  & Pruned Acc(\%)  & Acc Loss(\%)  & Pruned FLOPs  & Param Sp(\%) \tabularnewline
\hline 
\hline 
0.1  & 98.50  & 98.54$\pm$0.04  & -0.04  & 10.00  & 4.51 \tabularnewline
\hline 
0.2  & 98.50  & 98.63$\pm$0.08  & -0.13  & 15.70  & 7.57 \tabularnewline
\hline 
0.3  & 98.50  & 98.4$\pm$0.11  & 0.10  & 38.10  & 18.56 \tabularnewline
\hline 
0.4  & 98.50  & 98.46$\pm$0.18  & 0.04  & 32.20  & 20.34 \tabularnewline
\hline 
0.5  & 98.50  & 98.42$\pm$0.12  & 0.08  & 42.20  & 21.68\tabularnewline
\hline 
CGES  & 98.47  & 97.12  & 1.35  & 29.00  & 67.92\tabularnewline
\hline 
\end{tabular}\caption{\label{tab:Results-about-MNIST}Results about different pruning rate
performed in MNIST. In first row and second row, pruned network has
better performance because of more efficient learning. In third row,
pruning rate 0.4 is a good balance between accuracy and FLOPs}
\end{table}

\begin{table}
\begin{centering}
\begin{tabular}{c|c|c|c}
\hline 
 & Full model  & CGES()  & ours\tabularnewline
\hline 
\hline 
Accuracy(\%)  & 71.09  & 70.05  & 71.09 \tabularnewline
\hline 
Weight pruned(\%)  & -{}-  & 3.60  & 18.13\tabularnewline
\hline 
FLOPs pruned(\%)  & -{}-  & –  & 6.42\tabularnewline
\hline 
\end{tabular}
\par\end{centering}
\caption{\label{tab:On-LeNet-5}On LeNet-5, pruned weights and accuracy with
different methods.}
\end{table}

\begin{table}
\begin{centering}
\begin{tabular}{c|c|c}
\hline 
 & Full model  & Ours\tabularnewline
\hline 
\hline 
Accuracy(\%)  & 91.74 & 90.80 \tabularnewline
\hline 
Weight pruned(\%)  & -{}-  & 12.10\tabularnewline
\hline 
FLOPs pruned(\%)  & -{}-  & 11.3\tabularnewline
\hline 
\end{tabular}
\par\end{centering}
\caption{\label{tab:On-ResNet-32}On ResNet-32, pruned weights and accuracy with
different methods.}
\end{table}

\begin{figure}
\begin{centering}
\includegraphics[scale=0.6]{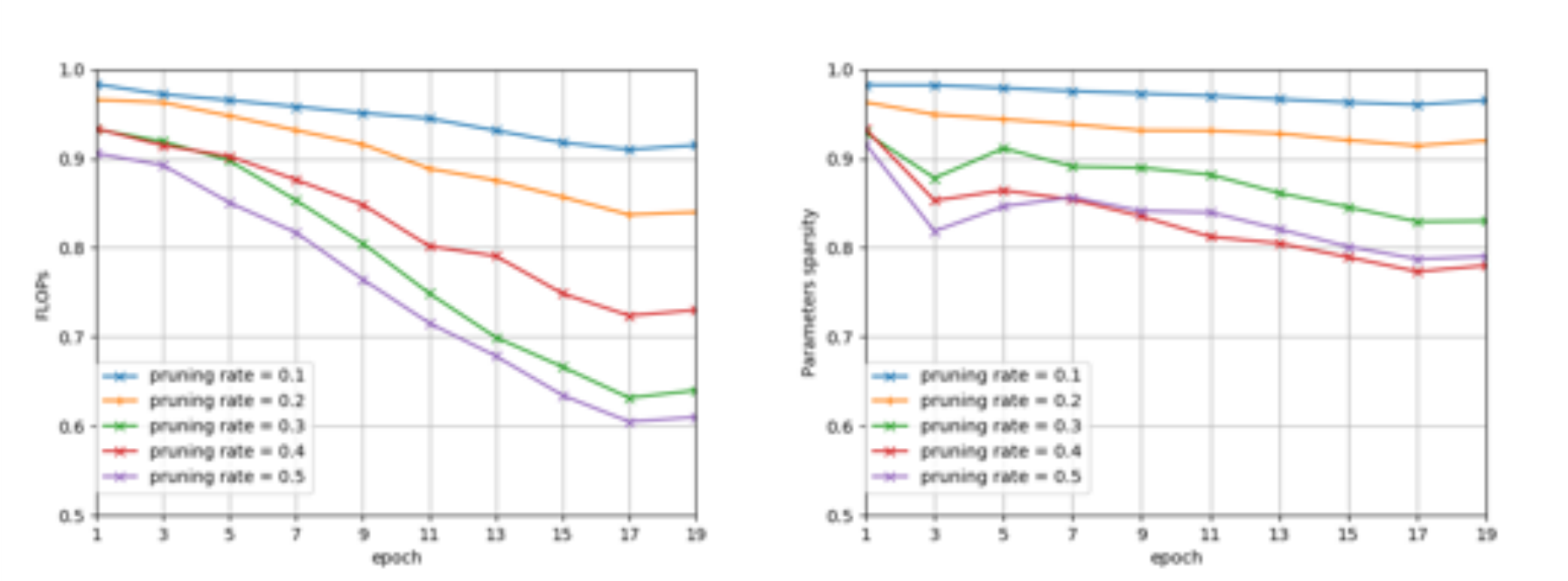} 
\par\end{centering}
\caption{\label{fig:FLOPs-and-parameters}FLOPs and parameters sparsity change
with training epochs. }
\end{figure}

\begin{table}
\centering{}%
\begin{tabular}{c|c|c|c|c}
\hline 
Layer  & Weights(K)  & FLOPs(K)  & Pruned Weights(\%)  & Pruned FLOPs(\%)\tabularnewline
\hline 
\hline 
conv1  & 0.38  & 587.94  & 53.13  & 53.13 \tabularnewline
\hline 
conv2  & 37.07  & 14,528.96  & 27.60  & 27.60 \tabularnewline
\hline 
fc1  & 2,483.91  & 4,967.83  & 22.65  & 22.65\tabularnewline
\hline 
total  & 2,599.52  & 18,812.688384  & 20.34  & 32.20\tabularnewline
\hline 
\end{tabular}\caption{\label{tab:Using-two-convolutional-MNIST}Using two convolutional
layers and two fully connected layers in MNIST.}
\end{table}

\subsection{Ablation Study}
\noindent \textbf{How the FLOPs and parameters sparsity changes during training?} In order to understand the process of pruning, we show the varying curve of FLOPs and parameters sparsity with five different pruning ratios in Figure \ref{fig:FLOPs-and-parameters}. It's intuitional that both FLOPs and parameters sparsity are decreasing progressively during training because some unimportant filters are removed. However, there are some fluctuations in curves, especially in the right one. This is because (1) we cluster all filters before pruning, and remove groups of filters if they are unimportant. Hence, the number of pruned filters depends on the situation of weightes updating; (2) we do filter pruning with soft self-adaption manners, which means the pruned filters will be updated iteratively during training. We argue that because of two reasons mentioned above, our proposed approach can achieve considerable performance with fewer parameters with fewer parameters.

\noindent \textbf{Which part of filters are more important in each layer?} As we explain above, if a group of filters is insensitive to the output, then it will be pruned, \emph{i.e.}, these filters are not important. In order to analyze the important of filtrs in each layer, we do experiments on MNIST with LeNet-4, as shown in Figure \ref{tab:Using-two-convolutional-MNIST}. Intertestingly, we can find that the more a layer is close to the output (the last fully-connected layer), the less filters are pruned. The superficial layers (\emph{e.g.}, conv1) only learn some simple features, such as color and edges, however, the deeper layers (\emph{e.g.}, fc1) can learn more abstract features, such as profile and parts. Therefore, it's intuitional that some parameters in fc1 are more important than those in conv1. Again, we argue that our proposed approach which apply spectral clustering in filter pruning is effective and explicable.

\section{Conclusion}

In this paper, we introduce a novel filter pruning method – spectral
clustering filter pruning with soft self-adaption manners(SCSP), to
compress the convolutional neural networks. We for the first time,
employ the spectral clustering on filters layer by layer to explore
their intrinsic connections. The experiments show the efficacy of
proposed methods.

{\small{}{}  \bibliographystyle{plain}
\bibliography{egbib}
 }{\small \par}

\end{document}